\documentclass[times,twocolumn,final,authoryear]{elsarticle}

\usepackage{prletters}

\usepackage{amssymb}
\usepackage{latexsym}

\usepackage{url}
\usepackage{xcolor}
\definecolor{newcolor}{rgb}{.8,.349,.1}

\usepackage{graphicx}
\usepackage{amsmath}
\usepackage{amssymb}
\usepackage{multirow}
\usepackage[draft]{hyperref}
\usepackage{threeparttable}
\usepackage{booktabs}

\journal{Pattern Recognition Letters}

\begin{document}

\thispagestyle{empty}

\clearpage
\thispagestyle{empty}
\ifpreprint
  \vspace*{-1pc}
\fi

\ifpreprint
  \setcounter{page}{1}
\else
  \setcounter{page}{1}
\fi

\begin{frontmatter}

\title{Unsupervised object-level video summarization with online motion auto-encoder}

\author[1]{Yujia \snm{Zhang}\corref{cor1}} 
\cortext[cor1]{Corresponding author. Work done while the first auther was at CMU.}
\ead{zhangyujia2014@ia.ac.cn}
\author[2]{Xiaodan \snm{Liang}}
\author[3]{Dingwen \snm{Zhang}}
\author[1]{Min \snm{Tan}}
\author[2]{Eric P. \snm{Xing}}

\address[1]{Institute of Automation, Chinese Academy of Sciences; University of Chinese Academy of Sciences, Beijing, China}
\address[2]{Carnegie Mellon University, Pittsburgh, PA, USA.}
\address[3]{Xidian university, Xi'an, China.}

\received{1 May 2013}
\finalform{10 May 2013}
\accepted{13 May 2013}
\availableonline{15 May 2013}
\communicated{S. Sarkar}

\begin{abstract}
Unsupervised video summarization plays an important role on digesting, browsing, and searching the ever-growing videos every day, and the underlying fine-grained semantic and motion information (i.e., objects of interest and their key motions) in online videos has been barely touched. In this paper, we investigate a pioneer research direction towards the fine-grained unsupervised object-level video summarization. It can be distinguished from existing pipelines in two aspects: extracting key motions of participated objects, and learning to summarize in an unsupervised and online manner. To achieve this goal, we propose a novel online motion Auto-Encoder (online motion-AE) framework that functions on the super-segmented object motion clips. Comprehensive experiments on a newly-collected surveillance dataset and public datasets have demonstrated the effectiveness of our proposed method.
\end{abstract}


\end{frontmatter}


\section{Introduction}
Video has rapidly become one of the most common sources of visual information. Automatic tools for analyzing and understanding video contents are essential for the large-scale intelligent system. In particular, automatic video content summarization techniques, e.g.,~\citep{gygli2014creating,potapov2014category}, have received wide research interest in recent years due to its huge application potentials for video analysis~\citep{chang2017semantic,chang2017feature,chang2017bi,ma2017many}. The goal is to compactly depict the original video, distilling its important events into a short watchable synopsis.

\begin{figure}[t]
\centerline{\includegraphics[width=8.5cm]{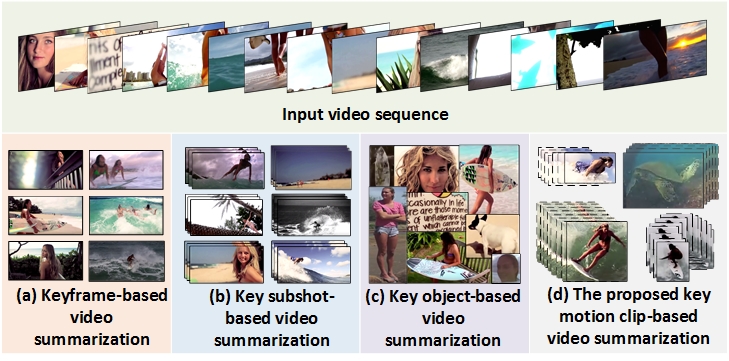}}
\vspace{-0.2cm}
\caption{Different types of video summarization techniques. Specifically, existing methods including (a) keyframe, (b) key subshot (c) key object-based (only spatial) video summarization. Our proposed unsupervised object-level (motion-clip based) video summarization technique (d) attempts to mine key object motion clips to compactly depict the whole video.}
\label{illustration_}
\vspace{-0.5cm}
\end{figure}

As shown in Fig.~\ref{illustration_} (a)$-$(c), the existing video summarization techniques often shorten the input video mainly in three different ways, i.e., keyframe selection, key subshot selection, key object selection within frames, respectively. Specifically, the keyframe-based approaches (e.g.~\citep{khosla2013large,kim2014joint}) generate the video summary comprised by a collection of key frames, while the key subshot-based approaches (e.g.~\citep{lu2013story,lin2015summarizing}) select the representative subshots of frames to form the video summary. Recently, the key object-based approach~\citep{meng2016keyframes} proposes to break down the whole video into several single frames that reveal the representative objects existing in a given video.

Despite the great progress achieved by the prior works, the underlying fine-grained semantic and motion information (i.e., objects of interest and their key motions) in online videos has been barely touched, which is more essential and beneficial for many down-streaming tasks (e.g., object retrieval) in an intelligent system. Besides, among the prior works, most of them (e.g.,~\citep{gong2014diverse,gygli2015video,zhang2016summary}) address the ill-posed supervised or semi-supervised video summarization with the requirement of predefining the patterns of key frames, which is impractical and not scalable to handle diverse and complicated ever-growing video contents. 

\begin{figure}[t]
\centerline{\includegraphics[width=8.5cm]{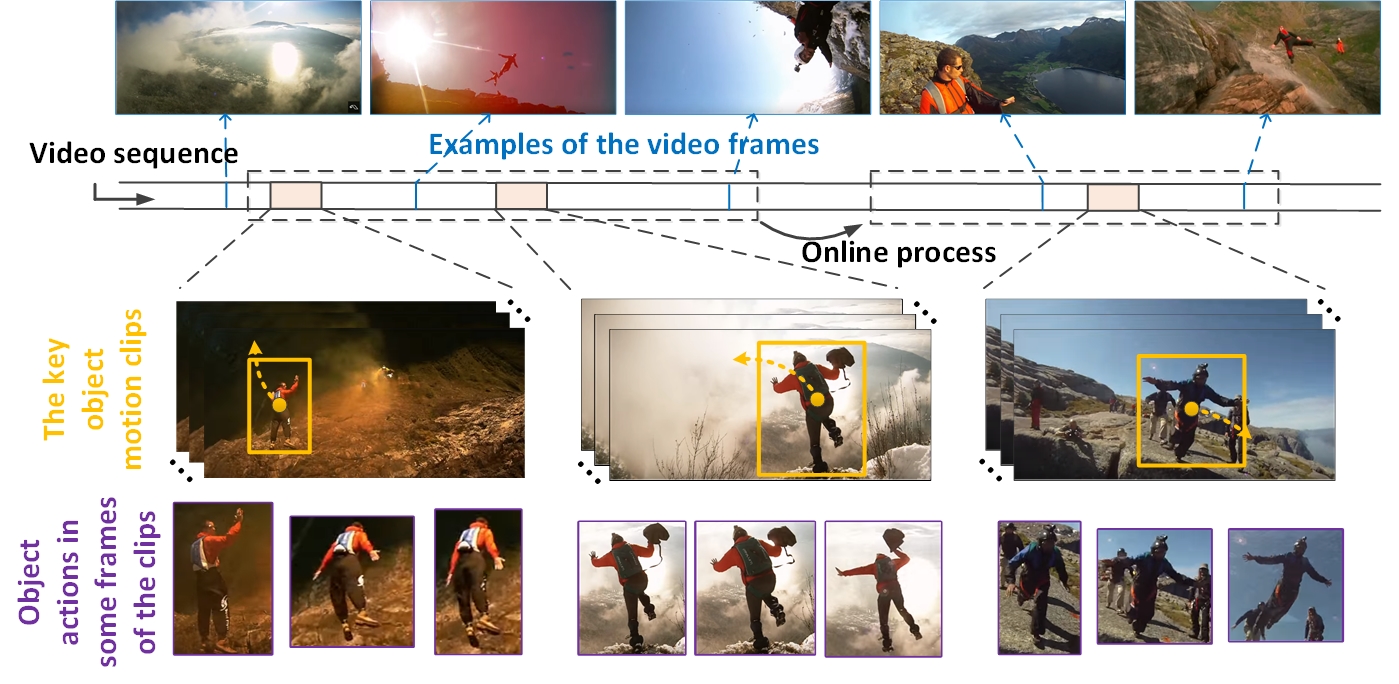}}
\vspace{-0.3cm}
\caption{The proposed unsupervised object-level video summarization attempts to mine key object motion clips to compactly depict the whole video. The dashed lines in the figure indicate the moving trajectories of the object instances in the extracted key object motion clips. Essentially, the video summaries generated by the proposed approach can not only answer the question of ``what are the representative objects residing in the video?'' but also answer ``what attractable actions of these objects are occurring in the video?''}
\label{illustration}
\vspace{-0.5cm}
\end{figure}

\begin{figure*}[t]
\centerline{\includegraphics[width=0.95\textwidth]{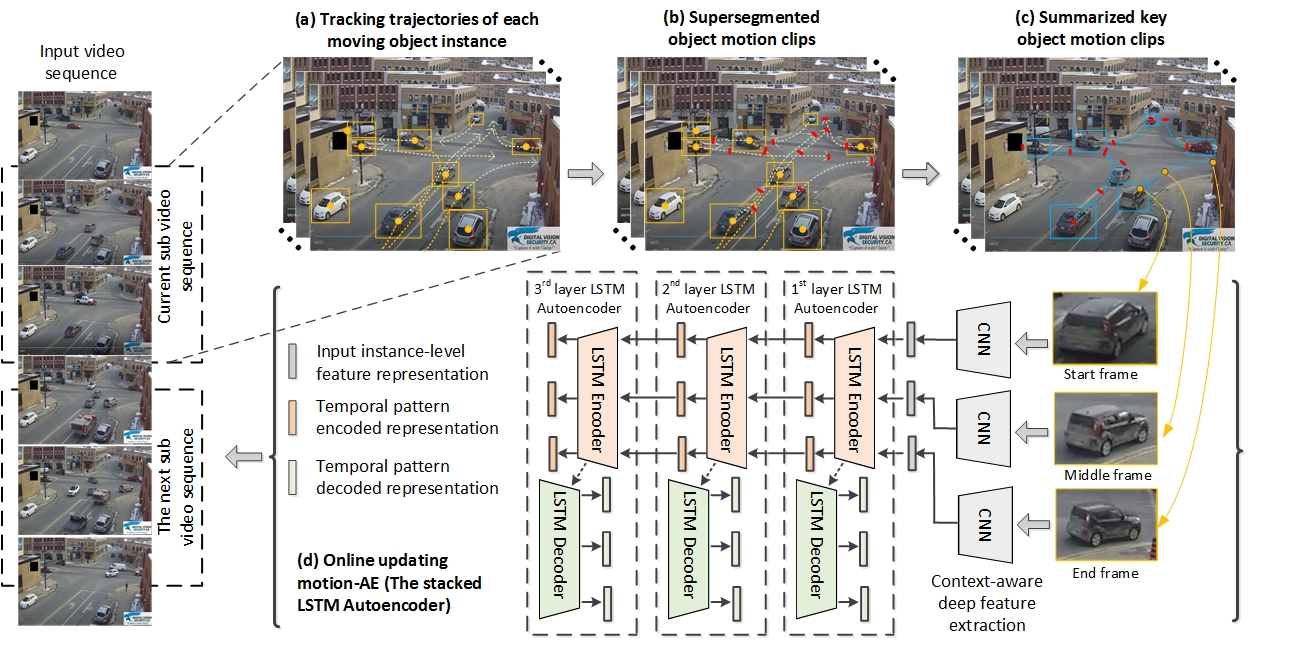}}
\vspace{-0.4cm}
\caption{The framework of the proposed online motion-AE model. Briefly, it sequentially reads and preprocesses each sub video sequence in an online manner. For each sub video sequence, we first track all the appearing object instances (as shown in (a)) and segment each of them into multiple object motion clips (as shown in (b)). Then, the deep LSTM auto-encoder (as shown in (d)) obtained previously is adopted to select the key object motion clips in the current video sequence. The output of deep LSTM auto-encoder is the reconstruction loss, which can be naturally used as the summarization score. Next, all the clips in the current video sequence are used to update the deep LSTM auto-encoder, which will be used in the next upcoming sub video sequence.  Notice that the dashed arrows indicate the motion trajectories of the object instance. Specifically, the yellow/green dashed arrows in (a) and (b) indicate the motion trajectories before/after the current video frame shown in the figure. The blue ones in (c) indicate the motion trajectories of the summarized key motion clips. The red dots in (a), (b), and (c) indicate the separation point of adjacent object motion clips.}
\label{framework}
\vspace{-0.3cm} 
\end{figure*}

To this end, this paper investigates a pioneer research direction towards the fine-grained unsupervised object-level video summarization as show in Fig.~\ref{illustration_} (d) and Fig.~\ref{illustration}. Such new video summarization can be distinguished from the existing ones mainly in two aspects: extracting key motions of participated objects and learning to summarize in an unsupervised and online manner that is more applicable for online growing videos. Essentially, the video summaries generated by the proposed approach can not only answer the question of ``what are the representative objects residing in the video?'' but also answer ``what attractable actions of these objects are occurring in the video?'' As is shown in Fig.~\ref{framework}, we propose a novel online motion Auto-Encoder (online motion-AE) framework. In summary, this paper makes the following four-fold contributions:

\begin{itemize}
\item \textbf{Key object motion-based video summarization.} We explore a pioneer research direction towards the fine-grained unsupervised video summarization that dives into the key object motion clips of a video stream to compactly depict the whole video and generate video summaries.

\item \textbf{Unsupervised online dictionary learning.}
We propose a novel online motion-AE model, which can mimic the online dictionary learning for memorizing past states of object motions by continuously updating a tailored recurrent auto-encoder network.

\item \textbf{The newly-collected OrangeVille benchmark.} A new surveillance video dataset is collected that allows for the objective evaluation of our new field of video summarization methods. We provide the spatial-temporal annotations for all key object motion clips to push forward the video summarization research with diverse granularities.

\item \textbf{State-of-the-art performance for both object motion-level and frame-level summarization.} Besides the key object motion-based summarization comparison on OrangeVille, we also conduct comprehensive experiments on other existing video summarization benchmarks to demonstrate the effectiveness of the proposed approach.
\end{itemize}

\section{Related Work}

There are mainly four lines of the existing video summarization techniques, i.e., the keyframe-based approach, the key subshot-based approach, the key object-based approach, and others, respectively. 

The keyframe-based video summarization methods~\citep{khosla2013large,kim2014joint,gong2014diverse} aim at identifying a series of discontinuous frames to form a summary that can represent the main video content well. For example, Khosla et al.~\citep{khosla2013large} used web-images as a prior to facilitate  video summarization. The intuition is that people tend to take pictures of objects and events from a few canonical viewpoints in order to capture them in a maximally informative way. Kim et al.~\citep{kim2014joint} achieved video summarization by diversity ranking on the similarity graphs between Flickr images and YouTube video frames. 

Different from the keyframe-based approaches, the key subshot-based video summarization approaches~\citep{lu2013story,yao2016highlight} aim at identifying a series of defining subshots, each of which is a temporally contiguous set of frames spanning a short time interval. For example, Song et al.~\citep{song2015tvsum} presented to use co-archetypal analysis technique to select shots which are most relevant to canonical visual concepts shared between video and images, according to the results of title-based image search.

More recently, Meng et al.~\citep{meng2016keyframes} proposed to summarize the video content into a collection of key objects, leading to the key object-based video summarization. Specifically, they formulated this representative selection problem as a sparse dictionary selection problem, i.e., choosing a few representative object proposals to reconstruct the whole proposal pool, then proposed to incorporate object proposal prior and locality prior in the feature space when selecting representative objects. The summarized key objects can be potentially used to facilitate the understanding of the video content in object level.

Besides the aforementioned video summarization manners, there are also some other interesting approaches, e.g,~\citep{sun2014salient,zhang2017revealing,chu2015video}. For example, Sun et al.~\citep{sun2014salient} proposed to summarize video content by first finding the ``montageable moments'' and then identifying salient people and actions to depict in each montage. Zhang et al.~\citep{zhang2017revealing} and Chu et al.~\citep{chu2015video} made attempt to summarize the related video content from a given video collection rather than a single video.

\section{Online Motion Auto-Encoder (online motion-AE)}

The proposed online motion-AE framework resolves the fine-grained unsupervised video summarization problem by training a stacked sparse LSTM auto-encoder in an online manner. In next sections, we first introduce the main stacked sparse LSTM auto-encoder module, and then present the details in building our online motion-AE framework.

\begin{figure*}[t]
\centerline{\includegraphics[width=\textwidth]{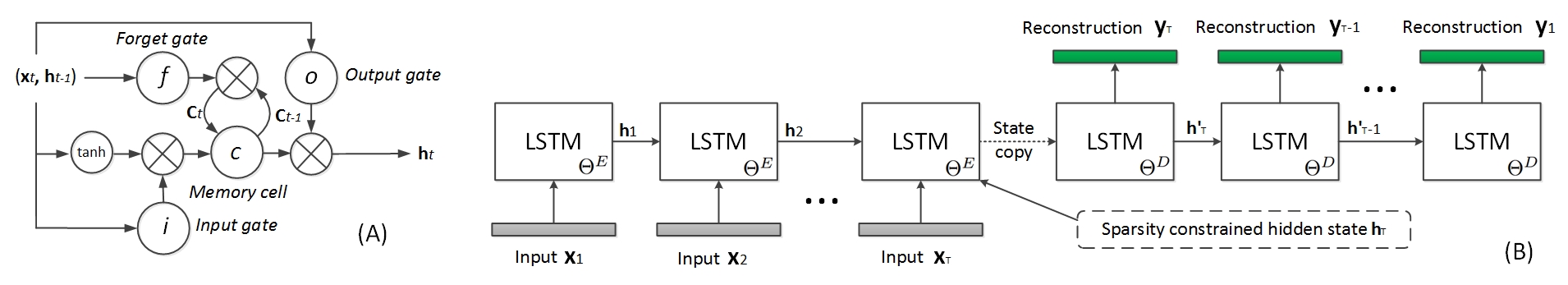}}
\vspace{-0.3cm}
\caption{The illustration of (A) the LSTM unit and (B) the sparse LSTM auto-encoder. LSTM unit is a type of recurrent neural network, which models long-range dependencies. Sparse LSTM auto-encoder is established by plugging the hierarchical three LSTM layers into a generative auto-encoder model with the sparse constraint.}
\label{LSTM}
\vspace{-0.5cm}
\end{figure*}

\subsection{The Stacked Sparse LSTM Auto-Encoder}

The key towards the fine-grained unsupervised object-level video summarization is the model capability of capturing key object-level motion clips within each video. Inspired by the success achieved by recurrent neural networks (especially Long Shot-Term Memory (LSTM)~\citep{hochreiter1997long}) on sequential modeling, our core learning module is established by plugging the hierarchical three LSTM layers into a generative auto-encoder model with the sparse constraint, as illustrated in Fig.~\ref{LSTM}. The sparse LSTM auto-encoder is thus able to learn the intrinsic motion patterns to reconstruct the input object sequence with the least possible amount of distortion. The learned parameters can be analogous to a compact ``motion dictionary'' of past observed object clips. Given a new object clip, we can easily judge its distinctness with respect to past ones by the reconstruction error induced from the learned LSTM auto-encoder model.

Specifically, the sparse LSTM auto-encoder model consists of a LSTM encoder and a LSTM decoder. The LSTM encoder recurrently takes each feature in the object clip as the input, and feed-forward it into the hierarchal LSTM layers at each time-step. After passing all features of the clip at all time-steps, a motion context vector is generated to encode the holistic temporal motion and appearance patterns. On the other hand, the LSTM decoder network decodes the extracted motion context vector by three symmetric LSTM layers to obtain a synthesized representation sequence. The sparse constraint is imposed on the encoded motion context vector to ensure the compactness and generalization capability of ``the motion dictionary". The reconstruction error between the synthesized features and input features is used as a measurement to determine whether this input sample has already been memorized by the recurrent auto-encoder. The unsupervised optimization of the sparse LSTM auto-encoder can be realized by the differential back-propagation algorithm to minimize the difference between the reconstructed features and input features. In the next, we will go through some details of the basic LSTM unit, the LSTM auto-encoder, the sparse LSTM auto-encoder, and the stacked sparse LSTM auto-encoder in order. 

\textbf{LSTM Unit.} As shown in Fig.~\ref{LSTM} (A), the basic LSTM unit, which is used for long-sequence modeling, applies a memory cell $\textbf{c}_t$ to record and encode the history of the knowledge of the inputs observed in previous time steps and determine the hidden cell in the current time step $\textbf{h}_t$. Here, the memory cell $\textbf{c}_t$ and hidden cell $\textbf{h}_t$ are modulated by nonlinear gate functions (e.g., the forget gate, output gate, and input gate), which determine whether the LSTM keeps the values at the gates (if the gates evaluate to 1) or discard them (if the gates evaluate to 0)~\citep{graves2014towards}.

Specifically, there are three types of gates: the input gate $\textbf{(i)}$ controls whether the LSTM considers its current input $\textbf{x}_t$, the forget gate $\textbf{(f)}$ allows the LSTM to forget its previous memory $\textbf{c}_t$, and the output gate $\textbf{(o)}$ decides how much of the memory to transfer to the hidden states $\textbf{h}_t$. These gates together lead to the ability on learning complex long-sequence modeling for LSTM. In particular, the forget gate $\textbf{(f)}$ serves as a time-varying data-dependent on/off switch to selectively incorporating the past and present knowledge~\citep{zhang2016video}.

The concrete formulations are defined as: 
  \begin{equation}\begin{split}
  &\textbf{i}_t = \sigma({W}_{ix}\textbf{x}_t+{\Phi}_{ih}\textbf{h}_{t-1}+\textbf{b}_i),\\
  &\textbf{f}_t = \sigma({W}_{fx}\textbf{x}_t+\Phi_{fh}\textbf{h}_{t-1}+\textbf{b}_f),\\
  &\textbf{o}_t = \sigma({W}_{ox}\textbf{x}_t+\Phi_{oh}\textbf{h}_{t-1}+\textbf{b}_o),\\
  &\textbf{c}_t = \textbf{i}_t \otimes \phi({W}_{cx}\textbf{x}_t+\Phi_{ch}\textbf{h}_{t-1}+\textbf{b}_c) + \textbf{f}_t \otimes \textbf{c}_{t-1},\\
  &\textbf{h}_t = \textbf{o}_t \otimes \phi(\textbf{c}_t),
  \end{split}\end{equation}
 where $\sigma$ is a sigmoid function, $\phi$ is the hyperbolic tangent function \textit{tanh}, $\otimes$  donates element-wise product, $\textbf{W} = \{{W}_{ix}, {W}_{fx}, {W}_{ox},{W}_{cx} \}$ is the transform from the input to LSTM states, $\mathbf{\Phi} = \{\Phi_{ih}, \Phi_{fh}, \Phi_{oh}, \Phi_{ch}\}$ is the recurrent transformation matrix between the hidden states, and $\textbf{b}$ is the bias vector.  

\textbf{LSTM Auto-Encoder.} 
In LSTM auto-encoder, both the encoder network and the decoder network are built upon the LSTM units. Given an input object clip $\textbf{X}=(\textbf{x}_1, \cdots , \textbf{x}_T)$, the LSTM encoder recurrently output hidden states $(\textbf{h}_1, \cdots , \textbf{h}_T)$ with shared network parameters $\Theta^{E}=\{\textbf{W}^{E},\mathbf{\Phi}^{E},\textbf{b}^{E}\}$, where $\textbf{W}^E$ is the transform from the input to LSTM encoder states. $\mathbf{\Phi}^E$ is the recurrent transformation matrix between the encoder states, and $\mathbf{b}^E$ is
the encoder bias vector. Symmetrically, the LSTM decoder recurrently decodes hidden states with the shared network parameters $\Theta^{D}=\{\textbf{W}^{D},\mathbf{\Phi}^{D},\textbf{b}^{D}\}$, and generates the current reconstruction output $\textbf{Y}=(\textbf{y}_1, \cdots , \textbf{y}_T)$ via an additional linear mapping:
 \begin{equation}
 \textbf{y}_t = {W}_{yh}\textbf{h}_t^\prime+\textbf{b}_h,
\end{equation}
where $\{{W}_{yh}, \textbf{b}_h\}$ are the reconstruction mapping parameters, and $\textbf{h}_t^\prime$ is the hidden state inferred for reconstructing the $t$-th input feature. 

Training the whole LSTM auto-encoder is to optimize the parameters $\{\Theta^{E},\Theta^{D},{W}_{yh}, \textbf{b}_h\}$ by minimizing the mean-squared reconstruction error between the input sequence and the corresponding reconstructed sequence via:
\begin{equation}
\underset{\Theta^{E},\Theta^{D},{W}_{yh}, \textbf{b}_h} {\arg \min} \mathcal{L}(\textbf{X},\textbf{Y}) = \frac{1}{2T} \sum_{t=1}^T ||\textbf{x}_t-\textbf{y}_t||_2^2.
\end{equation}

\textbf{Sparse LSTM Auto-Encoder.}
Motivated by the physiological evidence that describing patterns with less active neurons minimizes the probability of destructive cross-talk~\citep{olshausen2004sparse}, a regularization term is applied to constraining the sparsity to the target activation function~\citep{ng2011sparse,han2015background}. It penalizes a deviation of the expected activation of the hidden states representation from a fixed (low) level $\rho$ by optimizing:
\begin{equation}\begin{split}
&\underset{\Theta^{E},\Theta^{D},{W}_{yh}, \textbf{b}_h} {\arg \min} \sum_{k=1}^K \mathcal{L}(\textbf{X}^{(k)},\textbf{Y}^{(k)}) + \beta \sum_{d=1}^D KL(\rho||\widehat{\rho}_d),\\
&KL(\rho||\widehat{\rho}_d) = \rho \log \frac{\rho}{\widehat{\rho}_d} + (1-\rho) \log \frac{(1-\rho)}{(1-\widehat{\rho}_d)},
\label{objective function}
\end{split}\end{equation}
where $\textbf{X}^{(k)}$ indicates the $k$-th training sample in the training set of totally $K$ training sequences, $D$ indicates the dimension of the hidden states, $\rho$ is the target average activation of each dimension of the hidden state $\textbf{h}_T$, and $\widehat{\rho}_d = \sum_{k=1}^K [h_T^{(k)}]_d/K$ is the average activation of the $d$-th dimension of the hidden state $[h_T]_d$ over $K$ training samples. The Kullback-Leibler (KL) divergence $KL(\cdot)$ is used to impose the sparse constraint. As claimed in sparse coding~\citep{ng2011sparse}, a non-redundant over-complete feature set can be learned when $\rho$ is small. Here, we set $\rho = 0.05$ as suggested in~\citep{ng2011sparse}.

The architecture of a sparse LSTM auto-encoder network is shown in Fig.~\ref{LSTM} (B), which is established by plugging the hierarchical three LSTM layers into a generative auto-encoder model with sparse constraint. Notice that the output reconstruction sequence are in reverse order as compared with the input sequence as suggested by~\citep{srivastava2015unsupervised},  which could make the optimization easier because the model can get off the ground by looking at low range correlations.

\textbf{Stacked Sparse LSTM Auto-Encoder.} The final deep network is established by stacking $M = 3$ sparse LSTM auto-encoder layers as shown in Fig.~\ref{framework} (d). To achieve the hierarchical latent representation, the hidden states $(\textbf{h}_1,\cdots,\textbf{h}_T)$ from the first (bottom) LSTM auto-encoder layer is posed as the input $(\textbf{x}_1,\cdots,\textbf{x}_T)$ of the second LSTM auto-encoder layer. The dimension of hidden states in each LSTM layer is gradually decrease to capture more high-level motion representation by stacking more LSTM auto-encoder layers. Like many stacked auto-encoder models (e.g.,~\citep{srivastava2015unsupervised,han2015background}), the proposed deep LSTM auto-encoder model can be easily trained (both in the offline training stage and the online updating stage) using greedy optimization: training each of the sparse LSTM auto-encoder layer one by one via optimizing the objective function~\eqref{objective function} using stochastic gradient descent.

\subsection{Online Video Summarization}

Aiming at the fine-grained video summarization that dives into object instances, our online learning framework incorporates the multiple object tracking and motion trajectory segmentation to extract candidate object motion clips, and then processes these clips with the stacked sparse LSTM auto-encoder model in an online manner.

\subsubsection{Preprocessing}

\textbf{Multiple Object Tracking.}
\label{MOT}
We first track all moving objects in the video to discover all objects of interest to be summarized. Notice that the single object tracking (SOT) techniques as used in some previous video understanding systems~\citep{liang2017learning} cannot be readily used in our task as 1) they would miss many objects that are absence at the first frame of the video (or shot), and 2) they tend to be time-consuming when there are intensive objects appearing in the video (e.g., the surveillance video in our case).

Due to arbitrary number of objects that may appear in each frame, we adopt a state-of-the-art multiple object tracking algorithm, i.e., the Markov Decision Process (MDP) tracker~\citep{xiang2015learning}, which is a tracking-by-detection algorithm equipped with effective tracking model in handling the appearance/disappearance of any object instance in the video. 

It can benefit from the advantages of both offline and online learning. Particularly, the localization of moving objects is achieved by performing the pre-trained Faster RCNN~\citep{girshick2015fast} model on PASCAL VOC~\citep{pascal-voc-2012} dataset on down-sampled frames.

\begin{figure}[t]
\centerline{\includegraphics[width=9 cm]{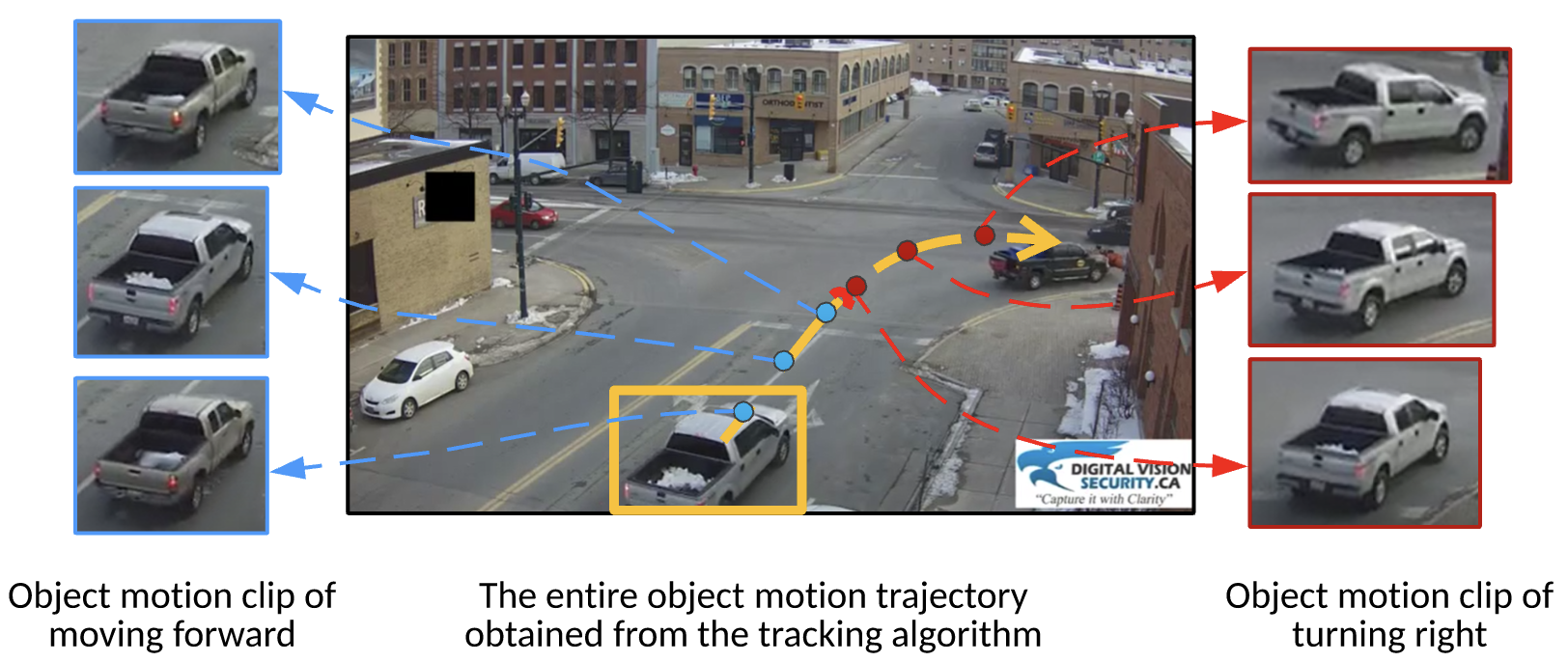}}
\vspace{-0.3cm}
\caption{An example of the motion trajectory segmentation. The motion trajectory of each object instance was super-segmented into multiple object motion clips, reflecting different motion status with unnoticeable changes.}
\label{superframe}
\vspace{-0.5cm}
\end{figure}

\begin{figure*}[t]
\centerline{\includegraphics[width=12cm]{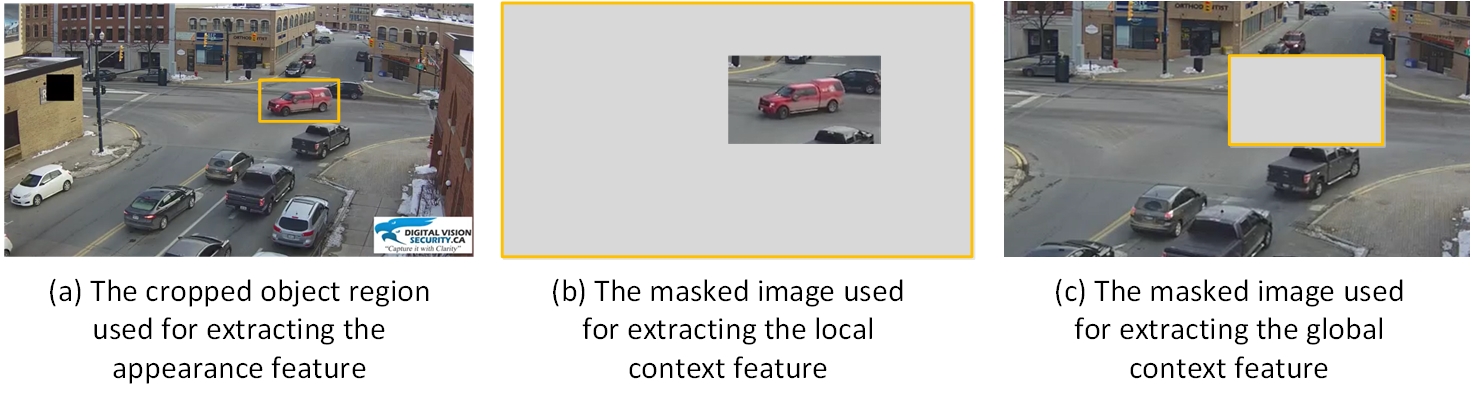}}
\vspace{-0.5cm}
\caption{Context-aware feature representation. The orange boxes are the bounding box locations feeding into the Faster RCNN to extract the corresponding features. (a) takes the object region as the input to generate the feature for the object; (b) and (c) are local and global context-aware feature representation respectively, where the local one is produced by replacing the pixel values out of the expanded bounding box for the object, while the global one is produced by replacing the pixel values inside of the expanded bounding box for the object.}
\centering
\label{features}
\vspace{-0.5cm}
\end{figure*}

\textbf{Motion Trajectory Segmentation.}
\label{MTS}
With the goal of finding the distinct motion patterns of each object, we thus super-segment the motion trajectory of each object instance into multiple object motion clips, each of which reflects a certain motion status with unnoticeable changes as shown in Fig.~\ref{superframe}. One straight-forward way would be to cut each motion trajectory into clips within a fixed length. However, such arbitrarily cutting clips would destroy some continuous motion sequences. An alternative way is to use the shot detection approach~\citep{smith1998video}, which is mainly based on changes in the color histogram. This still cannot work well in our case as the sequences formed by the cropped object trajectories usually contain the image regions with similar color distribution (see Fig.~\ref{superframe}). We thus adopt the recent superframe segmentation method~\citep{gygli2014creating}, which cuts video sequences mainly based on the motion information by defining an energy function.

\textbf{Context-aware Feature Representation.}
\label{SFE}
Given the super-segmented motion clips, we extract the context-aware feature representation for each region as the inputs to sparse LSTM auto-encoder. Specifically, we consider a key object motion clip from two prospectives: the distinct object appearance and distinguished surrounding context. We thus represent each region with a 8192-dimension vector by concatenating the appearance feature and the context feature extracted by the Faster RCNN network~\citep{li2016weakly}. Specifically, the appearance feature is the 4096-dimension feature vector generated by feeding the original object region into Faster RCNN network and using the ``fc7'' layer pooled from the corresponding object bounding box location (see Fig.~\ref{features} (a)). The context feature is another 4096-dimension feature vector generated by feeding the masked image, which is produced by replacing the pixel values out of the expanded bounding box\footnote{The bounding box is expanded to 2 times of its original length and width.} region with the fixed mean pixel values pre-computed on ILSVRC 2012~\citep{deng2009imagenet} into Faster RCNN network, and then using the ``fc7'' layer pooled from the entire image scene (see Fig.~\ref{features} (b)). This context-aware representation can not only capture the detailed appearance of each moving object, but also its corresponding location and surrounding context.

\subsubsection{Offline Training Process}
The sparse LSTM auto-encoder model is first initialized with an offline learning process to obtain the basic reconstruction capability, which is then updated in an online way. Specifically, we first collect a 40-minute-long (with 83278 frames) video sequence as the offline training data. Then, we randomly extract 30 bounding boxes from each offline video frame and copy each bounding box three times to form a still object sequence. In this way, the offline learned model will have the knowledge information for still object sequence, and thus can better differentiate moving object clips. Afterwards, the context-aware feature of each still sequence are thus used to train the stacked sparse LSTM auto-encoder.

\subsubsection{Online Summarization Process}

Given each test video sequence, the proposed online motion-AE sequentially reads the video 1000 frames by 1000 frames, (i.e., each of the 1000-frame sequence is treated as the sub video sequence as shown in Fig.~\ref{framework}) and summarizes the key object motion clips. At each step, we first extract candidate object motion clips, and sample three object motion regions (i.e., the start, middle, and end regions) from the clip as the input to the stacked sparse LSTM auto-encoder. This down-sampling operation help significantly reduces the computational complexity and yet retains the informative temporal information. We then extract the context-aware feature representation (as introduced in~\ref{SFE}) of the sampled object motion regions and use the pre-trained stacked sparse LSTM auto-encoder to obtain the reconstruction error of each object clip. Here the reconstruction errors can be naturally used as the summarization scores for unsupervised video summarization. Intuitively, key object motion clips with unseen appearance and motion cannot be well reconstructed from the patterns learned by either the offline video sequence or the previous video content in the online test video. Next, we use all clips in the current sub video sequence to update the stacked sparse LSTM auto-encoder for capturing the underlying motion patterns, and the updated network is employed on the upcoming sub video sequence. By performing the aforementioned online process for the whole video sequence, we can obtain the summarization score for each object motion clip, and thus get the final results for key object motion clip summaries.

\section{Experiment}
\subsection{Experimental Settings}
\textbf{Datasets.}
We evaluate the performance of the proposed online motion-AE model on several datasets. The first one is the newly collected video surveillance dataset named as OrangeVille, which contains 30 surveillance video sequences downloaded from the YouTube website with the keyword of ``OrangeVille'' and filtering out the videos with low resolution and strong shadows. Each of the video in OrangeVille dataset averagely contains 3000 frames with the length of 100 seconds.

This newly collected dataset we use mainly targets on fine-grained object-level video summarization task. Different from other keyframe-/key subshot-based datasets, the videos are recorded in outdoor scenes with moving objects mainly in categories of human, cars, buses, bicycles, and motorcycles. The motion/action of such objects are mainly \emph{waiting}, \emph{slow moving}, \emph{fast moving}, and \emph{turning around}. As this task aims to discover all key object motion clips, instead of labeling the key frames or key subshots as the other previous datasets did, we provide the spatial-temporal object bounding-box annotations to depict each key object motion clip in each video on OrangeVille dataset. The ground-truth annotations are labeled by a subject manually, focusing on the key object motions like fast moving cars, pedestrian crossing the road, and buses turning left/right.

The second dataset is the Base jumping dataset from public CoSum dataset~\citep{chu2015video}. Since CoSum dataset is applied for video co-summarization task, and the dataset is consisted of multiple videos organized into groups with topic keywords, we use one class of the whole dataset, i.e., the \emph{Base jumping} class, to evaluate our model. 

Besides, we also use other two frame-level datasets, which are the public SumMe dataset~\citep{gygli2014creating} and TVSum dataset~\citep{song2015tvsum}. Specifically, SumMe contains 25 videos covering \emph{holidays}, \emph{events} and \emph{sports}, such as Statue of Liberty, saving dolphins and bike polo. TVSum contains 50 videos in 10 categories downloaded from YouTube defined in the TRECVid Multimedia Event Detection (MED).

\textbf{Evaluation Metrics.}
We adopted three common metrics, including the average precision (AP) score, the F-measure score, and the AUC score, which have been used in previous works (e.g.,~\citep{zhang2016summary,gygli2015video} ). Specifically, The precision $PRE (Precision)$, true positive rate $TPR (Recall)$ and false positive rate $FPR$ were defined as:
  \begin{equation}\begin{split}
  & PRE=|SF \cap GF|/|SF|, \\
  & TPR=|SF \cap GF|/|GF|, \\
  & FPR=|SF \cap GB|/|GB|,
  \end{split}\end{equation}
where $SF$, $GF$ and $GB$ denote the set of samples predicted as 1, the samples labeled as 1, and the samples labeled as 0, respectively. F-measure was defined as:
 \begin{equation}
 F=\dfrac{2Precision\times Recall}{Precision + Recall}.
 \end{equation}
 The F-measure scores were obtained by binarizing the predictions via a threshold $\tau=0.5$.

In the fine-grained object-level video summarization task, the ground-truth are $G$ human labeled object motion clips $\{\textbf{f}_g,\textbf{b}_g^s,\textbf{b}_g^e\}_{g=1}^G$, where $\textbf{f}_g$,$\textbf{b}_g^s$, and $\textbf{b}_g^e$ indicate the temporal location (i.e., the frame numbers) of the $g$-th ground-truth clip, the spatial location (i.e., the object bounding box) in the start of the frame, and the spatial location in the end of the frame, respectively. Given an object motion clip $\{\textbf{f}_p,\textbf{b}_p^s,\textbf{b}_p^e\}$ which is predicted by the summarization algorithm, we calculate the IOU overlaps both for the spatial locations and the temporal locations as: 
\begin{equation}\begin{split}
&IOU^{tem}=\frac{\textbf{f}_p \cap \textbf{f}_g}{\textbf{f}_p \cup \textbf{f}_g},\\
&IOU^{spa}=\frac{1}{2} (\frac{\textbf{b}_p^s \cap \textbf{b}_g^s}{\textbf{b}_p^s \cup \textbf{b}_g^s} + \frac{\textbf{b}_p^e \cap \textbf{b}_g^e}{\textbf{g}_p^e \cup \textbf{b}_g^e}). 
\end{split}\end{equation}
Then, the object motion clips would be considered as correct summarization if $IOU^{tem}$ and $IOU^{spa}$ are simultaneously larger than the threshold $\varphi$. Due to the limited accuracy of the object tracking and superframe segmentation algorithms, we set $\varphi=0.1$ during our evaluation.

\begin{table*}[t]
\small
	\centering\renewcommand\arraystretch{1}
    \caption{Comparison of the fine-grained video summarization results on the~\textbf{OrangeVille} Dataset.}\label{OrangeVille result}
	\begin{tabular}{c|c|ccc|ccc|ccc}
    	\toprule

		&\textbf{\citep{zhao2014quasi}}	  & \multicolumn{3}{c|}{\textbf{\citep{ng2011sparse}}}  & \multicolumn{3}{c|}{\textbf{\citep{srivastava2015unsupervised}}}  & \multicolumn{3}{c}{\textbf{OURS}} \\
		
		&  &1 layer	&2 layers	&3 layers & 1 layer	&2 layers	&3 layers  &1 layer	&2 layers	&3 layers  \\
		
		\hline
		AUC score	&0.4252	&0.4512	&0.4040	&0.4354	&0.4829	&0.5413	&0.5680	&0.4970	&0.5598	&\textbf{0.5908} \\
		AP score	&0.1542	&0.1615	&0.1493	&0.1705	&0.2177	&0.2476	&0.2638	&0.2275	&0.2796	&\textbf{0.2850} \\
		F-measure	&0.1284	&0.1325	&0.1420	&0.1662	&0.2310	&0.2546	&0.2795	&0.2364	&0.2721	&\textbf{0.2901} \\
		\bottomrule
	\end{tabular}
	\vspace{-0.3cm}
\end{table*}

\begin{table*}[t]
\small
	\centering\renewcommand\arraystretch{1}
	\caption{Comparison of different masks which is used to extract context features on stacked sparse LSTM auto-encoder models.}\label{comparison of masks}
	\begin{tabular}{c|ccc|ccc}
    	\toprule
		& \multicolumn{3}{c|}{\textbf{mask = zero}} & \multicolumn{3}{c}{\textbf{mask = mean}}   \\
		&1 layer	&2 layers	&3 layers	&1 layer	&2 layers	&3 layers\\
		\hline
		AUC score	&0.4706	&0.5178	&0.5775	&0.4970	&0.5598	&\textbf{0.5908} \\
		AP score	&0.2008	&0.2166	&0.2420	&0.2275	&0.2796	&\textbf{0.2850} \\
		F-measure	&0.1841	&0.1886	&0.1996	&0.2364	&0.2721	&\textbf{0.2901} \\
		\bottomrule
	\end{tabular}
	\vspace{-0.3cm}
\end{table*}

\begin{figure*}[t]
\centerline{\includegraphics[width=0.88\textwidth]{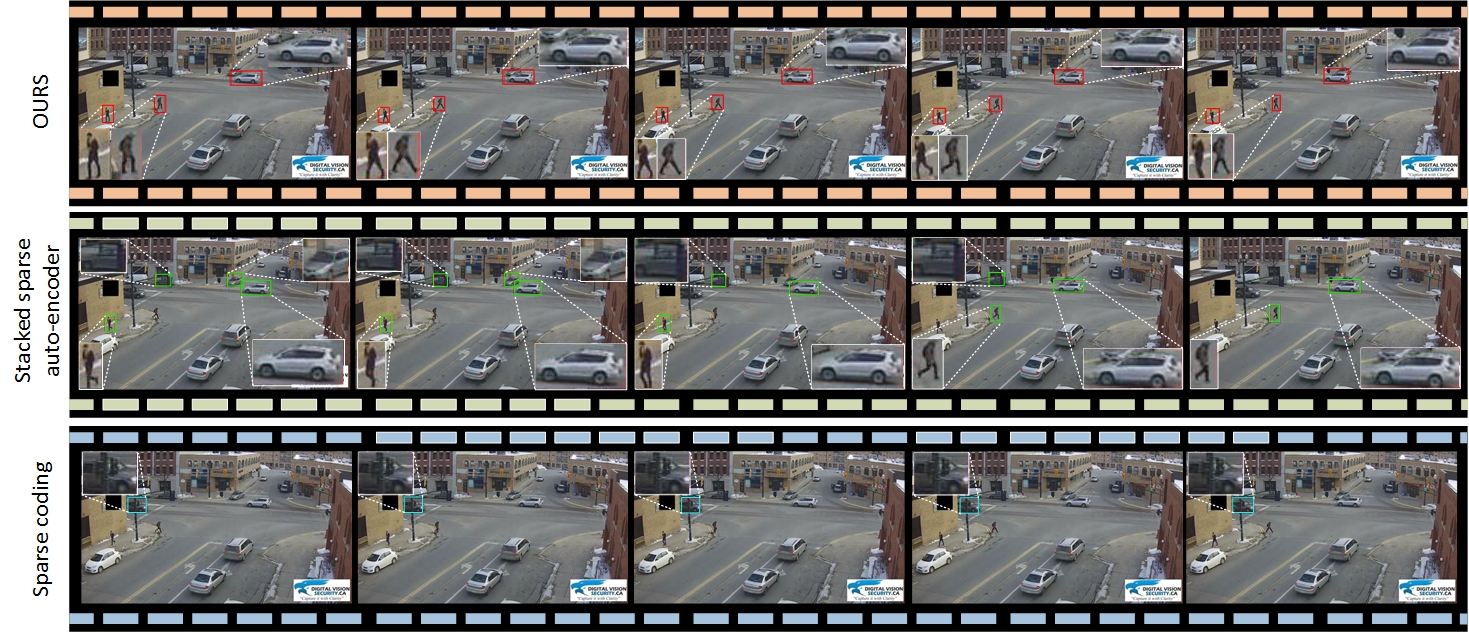}}
\vspace{-0.2cm}
\caption{Some examples of the fine-grained summarization results on the OrangeVille dataset. This shows the comparisons of our online motion-AE model with sparse coding~\citep{zhao2014quasi} and stacked sparse auto-encoder~\citep{ng2011sparse} models.
}
\label{fgres}
\vspace{-0.3cm}
\end{figure*}

\begin{table*}
\small
	\centering\renewcommand\arraystretch{1}
	\caption{Comparison of different features on stacked sparse LSTM auto-encoder models.}\label{comparison of features}
	\begin{tabular}{c|ccc|ccc|ccc}
		\toprule
		& \multicolumn{3}{c|}{\textbf{global + local}} & \multicolumn{3}{c|}{\textbf{object + global}} & \multicolumn{3}{c}{\textbf{object + local}}  \\
		&1 layer	&2 layers	&3 layers	&1 layer	&2 layers	&3 layers    &1 layer	&2 layers	&3 layers\\
		\hline
		AUC score	&0.4889	&0.4712	&0.5175    &0.4628	&0.4857	&0.5018	&0.4970	&0.5598	&\textbf{0.5908} \\
		AP score	&0.1928	&0.1934	&0.2026    &0.1677	&0.1834	&0.1864	&0.2275	&0.2796	&\textbf{0.2850} \\
		F-measure	&0.1839	&0.1764	&0.1362    &0.1050	&0.1216	&0.1404	&0.2364	&0.2721	&\textbf{0.2901} \\
		\bottomrule
	\end{tabular}
	\vspace{-0.3cm}
\end{table*}

\begin{table*}[t]
\small
	\centering\renewcommand\arraystretch{1}
	\caption{Comparison of the number of online updates on stacked sparse LSTM auto-encoder models.}\label{comparison of online updates}
	\begin{tabular}{c|ccc|ccc|ccc}
		\toprule
		& \multicolumn{3}{c|}{\textbf{1 online update}} & \multicolumn{3}{c|}{\textbf{2 online updates}} & \multicolumn{3}{c}{\textbf{3 online updates}}  \\
&1 layer	&2 layers	&3 layers	&1 layer	&2 layers	&3 layers    &1 layer	&2 layers	&3 layers\\
		\hline
		AUC score	&0.4427	&0.5359	&0.5826	&0.4970	&0.5598	&\textbf{0.5908}  &0.5268	&0.5491	&0.5710\\
		AP score	&0.2084	&0.2577	&0.2801	&0.2275	&0.2796	&\textbf{0.2850}  &0.2369	&0.2695	&0.2630\\
		F-measure	&0.2228	&0.2408	&0.2684	&0.2364	&0.2721	&\textbf{0.2901}  &0.2414	&0.2539	&0.2157\\
		\bottomrule
	\end{tabular}
	\vspace{-0.3cm}
\end{table*}

\begin{table*}[t]
\small
	\centering\renewcommand\arraystretch{1}
	\caption{Comparison of the number of layers on stacked sparse LSTM auto-encoder models.}\label{comparison of number of layers}
	\begin{tabular}{c|c|c|c|c|c}
		\toprule
		& {\textbf{1 layer}} & {\textbf{2 layers}} &{\textbf{3 layers}} & {\textbf{4 layers}} & {\textbf{5 layers}} \\
		\hline
	AUC score	&0.4970	&0.5598 &\textbf{0.5908}  &0.5919 &0.5875\\
	AP score	&0.2275	&0.2796 &\textbf{0.2850}  &0.2582	&0.2719\\
	F-measure	&0.2364	&0.2721 &\textbf{0.2901}  &0.2712	&0.2612\\
		\bottomrule
	\end{tabular}
	\vspace{-0.3cm}
\end{table*}

\begin{figure*}[t]
\centerline{\includegraphics[width=0.78\textwidth]{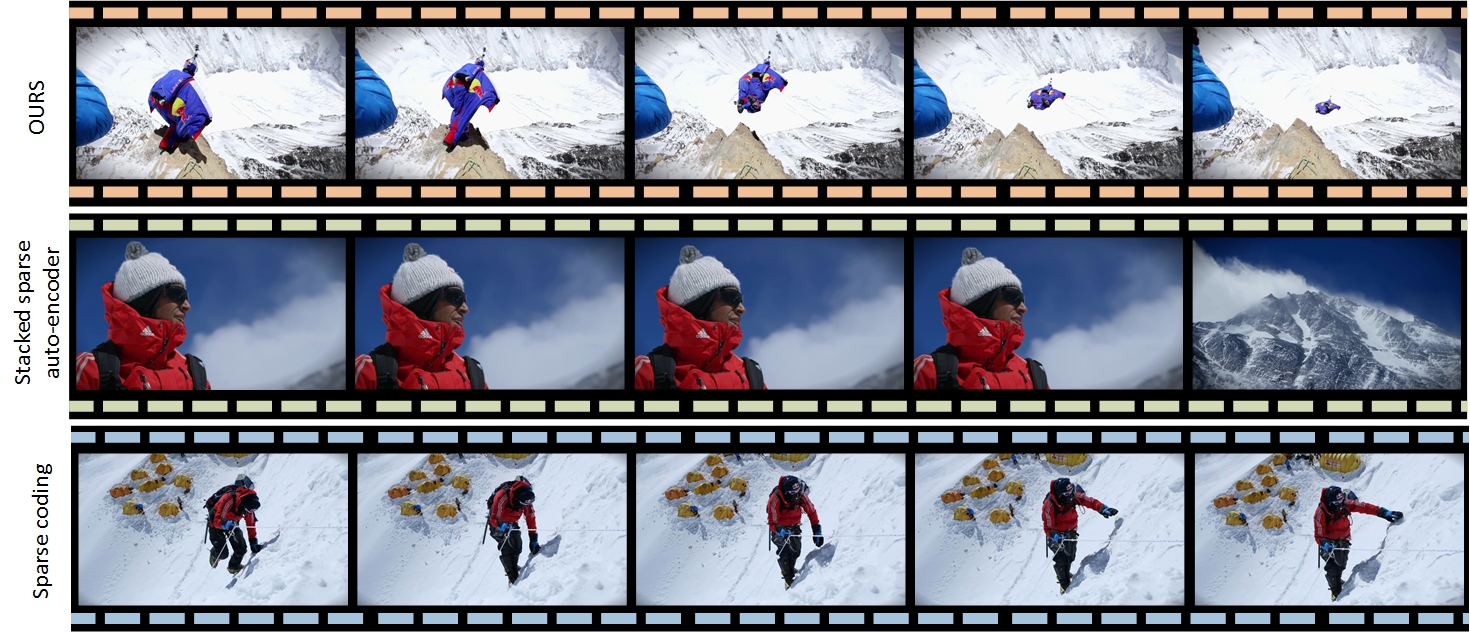}}
\vspace{-0.2cm}
\caption{Some examples of the frame-level summarization results on the Base jumping dataset. This shows the comparisons of our online motion-AE model with sparse coding~\citep{zhao2014quasi} and stacked sparse auto-encoder~\citep{ng2011sparse} models.}
\centering
\label{superres}
\vspace{-0.4cm}
\end{figure*}

\textbf{Implementation Details.} We implemented our model on the public TensorFlow platform. The dimensions of hidden states of the first, second, and third sparse LSTM auto-encoder layers were 4096, 2048, and 1024, respectively. In the offline pre-training phase, we set the batch size as 100, while in the online updating phase, the batch size was dynamically determined by the number of clips extracted in each sub video sequence. The learning rate was set to 0.001 during the offline pre-training while 0.0001 during the online updating.

\subsection{Fine-Grained Video Summarization}
In this section, we conducted extensive evaluation on OrangeVille dataset.

\textbf{Comparison with state-of-the-art and baseline learning models.}
To demonstrate the superiority of the proposed online motion-AE, we conducted comparisons with two state-of-arts of the learning models, which are the online sparse coding as used in~\citep{zhao2014quasi} and the stacked sparse auto-encoder as introduced in~\citep{ng2011sparse}, respectively. Particularly, the online sparse coding proposed in~\citep{zhao2014quasi} is one of the current state-of-the-art model for frame-level video summarization both in the online and unsupervised fashion. Other state-of-the-art video summarization approaches cannot be readily used in our comparison as they do not work in the unsupervised and online manner.

As for the online sparse coding-based approach, we firstly used the extracted object motion clips in the offline video sequence to learn a dictionary. Then, in the online testing phase, we used the previously constructed dictionary to calculate the reconstruction error for each new object motion clip in the current sub video sequence. We adopted the same motion-trajectory segmentation and context-aware feature representation modules used in our model to enable a fair comparison of different learning approaches. As for the  state-of-the-art model of the stacked sparse auto-encoder, it can be implemented by simply replacing the LSTM units with a single non-linear layer in our model. 

The qualitative evaluation results are shown in Fig.~\ref{fgres}, from which we can observe that the proposed online motion-AE can effectively summarize the key object motion clips, e.g., the fast driving cars, the occasionally appeared walkers who are crossing the road and turning around in the surveillance video sequence. Whereas the stacked sparse auto-encoder and sparse coding-based approaches are more likely to select non-key object motion clips, e.g., the stationary cars waiting for the traffic lights.

The quantitative evaluation results are shown in Table~\ref{OrangeVille result}. We can observe that the auto-encoder based models can significantly outperform the online sparse coding methods, even with a single layer, which demonstrates the good reconstruction capability of auto-encoders.

We also made comparisons with several model variants to show the effectiveness of each model component: 1) the stacked LSTM auto-encoder that eliminates the sparsity on the hidden states in each layer; 2) the shallow LSTM auto-encoder with or without using the sparse constraint, which only adopts one single LSTM auto-encoder layer to generate reconstruction errors; 3) the two-layer LSTM auto-encoder models with or without the sparse constraint. From the obtained experimental results, it can be observed that using hierarchical LSTM auto-encoder structure as our full model can achieve better performance than the other variants that only adopt one or two layers. This is due to the more powerful hierarchical feature representations learned by our model, which can benefit for capturing complex object motion patterns. 

\begin{figure}[t]
\centerline{\includegraphics[width=6.2cm]{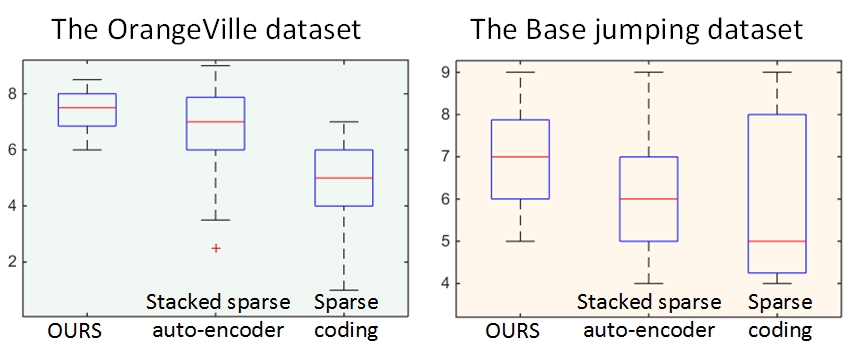}}
\vspace{-0.4cm}
\caption{The boxplot of the user study results on the collected OrangeVille dataset and the public Base jumping dataset on our online motion-AE model, stacked sparse auto-encoder~\citep{ng2011sparse} and sparse coding~\citep{zhao2014quasi} models.}
\label{plot}
\vspace{-0.4cm}
\end{figure}

\begin{table}[t]
\small
	\centering\renewcommand\arraystretch{1}
    \caption{Comparison of \textbf{\emph{Unsupervised}} frame-level video summarization results on \textbf{SumMe} Dataset.}\label{frama_level_comparisons}
	\begin{tabular}{c|c}
    	\toprule
		\textbf{Method} &\textbf{F\_measure}  \\
		\hline
		Video MMR~\citep{li2010multi}	 &0.266 \\
		TVSum~\citep{song2015tvsum}	 &0.266 \\
		VSUMM$_{1}$~\citep{de2011vsumm}	 &0.328 \\
		VSUMM$_{2}$~\citep{de2011vsumm}	 &0.337 \\
		stacked GRU Auto-Encoder~\citep{cho2014learning}	 &0.354 \\
		\textbf{Online Motion-AE}	&\textbf{0.377} \\
		\bottomrule
	\end{tabular}
    \label{fr_SumMe}
    \vspace{-0.2cm}
\end{table}

\begin{table}[t]
\small
	\centering\renewcommand\arraystretch{1}
    \caption{Comparison of \textbf{\emph{Unsupervised}} frame-level video summarization results on \textbf{TVSum} Dataset.}\label{frama_level_comparisons}
	\begin{tabular}{c|c}
    	\toprule
		\textbf{Method} &\textbf{F\_measure}  \\
		\hline
		Web Image Prior~\citep{khosla2013large}	 &0.360 \\
		LiveLight~\citep{zhao2014quasi}	 &0.460 \\
		TVSum~\citep{song2015tvsum}	 &0.500 \\
		stacked GRU Auto-Encoder~\citep{cho2014learning}	 &0.510 \\
		\textbf{Online Motion-AE} &\textbf{0.515} \\
		\bottomrule
	\end{tabular}
    \label{fr_TVSum}
    \vspace{-0.5cm}
\end{table}

\textbf{Ablation analysis.} We discuss different components in our stacked sparse LSTM auto-encoder model: 

1) the mask we used to extract context features, one is produced by replacing the pixel values out of the expanded bounding box region with the fixed mean pixel values which are pre-computed on ILSVRC 2012~\citep{li2016weakly}, and the other uses value of zero to replace the original pixel values. As shown in Table~\ref{comparison of masks}, we can see that using mask of mean pixel value performs better than mask of zero value in all layers;

2) the different types of features we used for object motion clip feature representation, including the concatenating the global context and local context features (\emph{global+local}), the appearance feature and global context feature (\emph{object+global}), and the appearance feature and local context feature (\emph{object+local}). The global and local context feature are illustrated in Fig.~\ref{features} (b)(c). The results are shown in Table~\ref{comparison of features}, and we can observe that local context-aware feature have better representation ability, due to the reason that local context preserves and highlights the moving regions while global context would weaken motion information for the moving objects;

3) the number of online updates we used to update during testing period. From Table~\ref{comparison of online updates}, we can see that using 1 and 2 online updates, the performance increases as the layers increase, but the performance decreases for 3 online updates due to the reason of over-learning;

4) the number of layers we used in our model. The results in Table~\ref{comparison of number of layers} demonstrate that the stacked sparse LSTM auto-encoder often converges in the 3-rd layer and then would slightly decline with 4, 5 and more layers.

\subsection{Frame-level Video Summarization}

We demonstrate the generalization capability of our model on frame-level video summarization task, which is evaluated on the Base jumping, SumMe and TVSum datasets.

\textbf{Our tailored pipeline for frame-level summarization on \emph{Base jumping} dataset.}
We adapt our model for frame-level summarization, where the summarization scores of a certain frame was obtained by averaging the reconstruction errors of all object motion clips residing in it.

The qualitative evaluation results are shown in Fig.~\ref{superres}, from which we can see that the proposed online motion-AE can summarize the key superframes that contain representative object motion, e.g., the moment for the jumping and flying. Whereas the stacked sparse auto-encoder and sparse coding-based approaches are more likely to select superframes with less informative object motion. 

\textbf{Our tailored pipeline for frame-level summarization on \emph{SumMe} and \emph{TVSum} datasets.}
We adapt our model for frame-level summarization, where the summarization scores of a certain superframe was obtained first by adopting the recent superframe segmentation method~\citep{gygli2014creating} to super-segment each video into multiple superframes, which cuts video sequences mainly based on the motion information. Then we used pre-trained ResNet152~\citep{he2016deep} model on ILSVRC 2015~\citep{ILSVRC15} to extract superframe motion features. After that, we used our online motion-AE model to obtain the reconstruction errors of all frame-level motion clips. In this way, the proposed approach as well as the baseline methods can be conveniently extended to obtain the frame-level summarization results.

\textbf{Comparison with competitive learning models.} The quantitative evaluation results for SumMe and TVSum datasets are shown in Table~\ref{fr_SumMe} and Table~\ref{fr_TVSum} respectively. The state-of-arts we choose are all unsupervised methods, which are the same of our experimental settings for fair comparisons. As can be seen, the proposed stacked sparse LSTM auto-encoder again obtains better performance than other unsupervised baselines on two different datasets. This demonstrates that the proposed learning model can effectively capture the informative temporal motion patterns and thus obtain the outperforming performance. The performance gaps among different approaches become smaller as compared with those in the fine-grained video summarization task. The reason is that the fine-grained video summarization requires the model be able to capture detailed instance-level motion patterns, which is much more challenging than the frame-level summarization task. 

\textbf{Underlying connections between the two tasks.} Compared with the key object-motion clips summarized by the fine-grained object-level video summarization, the key frames summarized by the frame-level video summarization consider not only the distinct foreground object motions but also the attractable whole image scenes. Thus, generating the frame-level summaries by only mining distinct object motion may partially cover the desired summarization result, which explains the performance gaps drop on the SumMe and TVSum datasets. We can observe that the performance drop of the proposed online motion-AE is not quite significant, which indicates that key object motions may actually occupy a relatively large portion of the factors in selecting frame-level summaries. Consequently, this demonstrates the potential usage of the fine-grained object-level video summarization approach.

\subsection{User Study}
We further provide human evaluation results on OrangeVille and Base jumping datasets to better verify the model capability. Similar with~\citep{khosla2013large}, 15 subjects are asked to watch the randomly selected videos at 3x speed and then showed them the corresponding video summaries generated by different approaches. They were asked to rate the overall quality of each summarization result by assigning a rating from 1 to 10 (higher is better). The user study results on the OrangeVille dataset are reported in Fig.~\ref{plot}. Consistent with the objective evaluation results obtained by the evaluation metrics, our approach can consistently outperforms the stacked sparse auto-encoder and spare coding models on the fine-grained object-level video summarization task. In addition, from the user study results, it can be seen that the proposed model can obtain more consistent subjective evaluation scores across different subject judges than the competitive models. Thus, this experiment further verifies the effectiveness of the proposed online motion-AE model.

\section{Conclusion}

In this paper, we investigated a pioneer research direction towards the unsupervised object-level video summarization. By diving into the fine-grained moving object instances residing in each given video sequence, the proposed online motion-AE approach can learn to summarize the extracted object motion clips in an unsupervised and online manner. Specifically, the online motion-AE can mimic the online dictionary learning for memorizing past states of object motions by continuously updating a tailored recurrent auto-encoder network, which enables the jointly online feature learning and dictionary learning to discriminate key motion clips. Comprehensive experiments on a newly collected OrangeVille dataset as well as public Base jumping, SumMe and TVSum datasets have demonstrated the effectiveness of the proposed approach both in the fine-grained video summarization and its application potential in the frame-level video summarization.

\section*{Acknowledgment}
This project is supported by the Department of Defense under Contract No. FA8702-15-D-0002 with Carnegie Mellon University for the operation of the Software Engineering Institute, a federally funded research and development center. This work is also supported by the National Natural Science Foundation of China (Grant No. 61673378 and 61333016).

\bibliographystyle{model2-names}
\bibliography{main.bbl}

\end{document}